\documentclass[12pt]{article}

\usepackage{arxiv}

\usepackage[utf8]{inputenc} 
\usepackage[T1]{fontenc}    
\usepackage{hyperref}       
\usepackage{url}            
\usepackage{booktabs}       
\usepackage{amsfonts}       
\usepackage{amsmath}
\usepackage{nicefrac}       
\usepackage{microtype}      
\usepackage{lipsum}
\usepackage{tikz}

\usepackage[authoryear]{natbib}

\usepackage{epsf}

\usepackage{psfrag}
\usepackage{graphicx}
\usepackage{amssymb}
\usepackage{amsbsy}
\usepackage{epsfig}
\usepackage{ifthen}
\usepackage{eucal}
\usepackage{theorem}

\newcommand{\commentout}[1]{}

\newcommand{\RR}{\mathbb{R}}    
\newcommand{\R}{\mathbb{R}}    

\newcommand{\Fcal}{\mathcal{F}} 

\newcommand{\x}{\mathbf{x}}    
\newcommand{\eq}[1]{(\protect\ref{#1})}

\newcommand{\independent}{\perp\mkern-11mu\perp}

\newcommand{\E}{\mathbb{E}}

\newcommand{\B}[1]{\mathbf{#1}}

\newcommand{\PA}{\B{PA}}

  \mathchardef\ordinarycolon\mathcode`\:
  \mathcode`\:=\string"8000
  \begingroup \catcode`\:=\active
    \gdef:{\mathrel{\mathop\ordinarycolon}}
  \endgroup

\title{Causality for Machine Learning}

\author{
  Bernhard Schölkopf
\\
  Max Planck Institute for Intelligent Systems, Max-Planck-Ring 4, 72076 T\"ubingen, Germany \\
  \texttt{bs@tuebingen.mpg.de} \\
}

\begin{document}
\maketitle

\begin{abstract}

  Graphical causal inference as pioneered by Judea Pearl arose from research on artificial intelligence (AI), and for a long time had little connection to the field of machine learning.
  This article discusses where links have been and should be established, introducing key concepts along the way. It argues that the hard open problems of machine learning and AI are intrinsically related to causality, and explains how the field is beginning to understand them.
\end{abstract}


\section{Introduction}
The machine learning community's interest in causality has significantly increased in recent years. My understanding of causality has been shaped by Judea Pearl and a number of collaborators and colleagues, and much of it went into a book written with Dominik Janzing and Jonas Peters \citep{PetJanSch17}. I have spoken about this topic on various occasions,\footnote{e.g., \citep{Schoelkopf2017icml}, talks at ICLR, ACML, and in machine learning labs that have meanwhile developed an interest in causality (e.g., DeepMind); much of the present paper is essentially a written out version of these talks} and some of it is in the process of entering the machine learning mainstream, in particular the view that causal modeling can lead to more invariant or robust models. There is excitement about developments at the interface of causality and machine learning, and the present article tries to put my thoughts into writing and draw a bigger picture. I hope it may not only be useful by discussing the importance of causal thinking for AI, but it can also serve as an introduction to some relevant concepts of graphical or structural causal models for a machine learning audience.

In spite of all recent successes, if we compare what machine learning can do to what animals accomplish, we observe that the former is rather bad at some crucial feats where animals excel. This includes transfer to new problems, and any form of generalization that is not from one data point to the next one (sampled from the same distribution), but rather from one problem to the next one –--- both have been termed {\em generalization}, but the latter is a much harder form thereof. This shortcoming is not too surprizing, since machine learning often disregards information that animals use heavily: interventions in the world, domain shifts, temporal structure –-- by and large, we consider these factors a nuisance and try to engineer them away. Finally, machine learning is also bad at {\em thinking} in the sense of Konrad Lorenz, i.e., acting in an imagined space. I will argue that causality, with its focus on modeling and reasoning about interventions, can make a substantial contribution towards understanding and resolving these issues and thus take the field to the next level. I will do so mostly in non-technical language, for many of the difficulties of this field are of a conceptual nature.

\section{The Mechanization of Information Processing}
The first industrial revolution began in the late 18th century and was triggered by the steam engine and water power.
The second one started about a century later and was driven by electrification. If we think about it broadly, then both are about how to generate and convert forms of {\bf energy}. Here, the word ``generate'' is used in a colloquial sense --- in physics, energy is a conserved quantity and can thus not be created, but only converted or harvested from other energy forms. 
Some think we are now in the middle of another revolution, called the digital revolution, the big data revolution, and more recently the AI revolution. The transformation, however, really started already in the mid 20th century under the name of cybernetics. It replaced energy by {\bf information}. Like energy, information can be processed by people, but to do it at an industrial scale, we needed to invent computers, and to do it intelligently, we now use AI. Just like energy, information may actually be a conserved quantity, and we can probably only ever convert and process it, rather than generating it from thin air. 
When machine learning is applied in industry, we often convert user data into predictions about future user behavior and thus money. Money may ultimately be a form of information –-- a view not inconsistent with the idea of bitcoins generated by solving cryptographic problems. The first industrial revolutions rendered energy a universal currency \citep{Smil17}; the same may be happening to information.

Like for the energy revolution, one can argue that the present revolution has two components: the first one built on the advent of electronic computers, the development of high level programming languages, and the birth of the field of computer science, engendered by the vision to create AI by manipulation of symbols. The second one, which we are currently experiencing, relies upon learning. It allows to extract information also from unstructured data, and it automatically infers rules from data rather than relying on humans to conceive of and program these rules. While Judea's approach arose out of classic AI, he was also one of the first to recognize some of the limitations of hard rules programmed by humans, and thus led the way in marrying classic AI with probability theory \citep{Pearl1988book}. This gave birth to graphical models, which were adopted by the machine learning community, yet largely without paying heed to their causal semantics. In recent years, genuine connections between machine learning and causality have emerged, and we will argue that these connections are crucial if we want to make progress on the major open problems of AI.

At the time, the invention of automatic means of processing energy transformed the world. It made human labor redundant in some fields, and it spawned new jobs and and markets in others. The first industrial revolution created industries around coal, the second one around electricity. The first part of the information revolution built on this to create electronic computing and the IT industry, and the second part is transforming IT companies into ``AI first'' as well as creating an industry around data collection and ``clickwork.'' While the latter provides labelled data for the current workhorse of AI, supervised machine learning \citep{Vapnik98}, one may anticipate that new markets and industries will emerge for causal forms of directed or interventional information, as opposed to just statistical dependences.


The analogy between energy and information is compelling, but our present understanding of information is rather incomplete, as was the concept of energy during the course of the first two industrial revolutions. The profound modern understanding of the concept of energy came with the mathematician Emmy Noether, who understood that energy conservation is due to a symmetry (or covariance) of the fundamental laws of physics: they look the same no matter how we shift the time, in present, past, and future. Einstein, too, was relying on covariance principles when he established the equivalence between energy and mass. 
Among fundamental physicists, it is widely held that information should also be a conserved quantity, although this brings about certain conundra especially in cosmology.\footnote{What happens when information falls into a black hole? According to the \emph{no hair conjecture}, a black hole seen from the outside is fully characterized by its mass, (angular) momentum, and charge.}
One could speculate that the conservation of information might also be a consequence of symmetries --- this would be most intriguing, and it would help us understand how different forms of (phenomenological) information relate to each other, and define a unified concept of information. We will below introduce a form of invariance/independence that may be able to play a role in this respect.\footnote{Mass seemingly played two fundamentally different roles (inertia and gravitation) until Einstein furnished a deeper connection in general relativity. It is noteworthy that causality introduces a layer of complexity underlying the symmetric notion of statistical mutual information. Discussing source coding and channel coding, \citet{Shannon59} remarked: {\em This duality can be pursued further and is related to a duality between past and future and the notions of control and knowledge. Thus we may have knowledge of the past but cannot control it; we may control the future but have no knowledge of it.}
  According to Kierkegaard, {\em Life can only be understood backwards; but it must be lived forwards.} }
The intriguing idea of starting from symmetry transformations and defining objects by their behavior under these transformations has been fruitful not just in physics but also in mathematics \citep{klein1872vergleichende,maclane:71}.

Clearly, digital goods are different from physical goods in some respects, and the same holds true for information and energy. A purely digital good can be copied at essentially zero cost \citep{NBERw25695}, unless we move into the quantum world \citep{1982Natur.299..802W}.
The cost of copying a physical good, on the other hand, can be as high as the cost of the original (e.g., for a piece of gold). In other cases, where the physical good has a nontrivial informational structure (e.g., a complex machine), copying it may be cheaper than the original.
In the first phase of the current information revolution, copying was possible for software, and the industry invested significant effort in preventing this. In the second phase, copying extends to datasets, for given the right machine learning algorithm and computational resources, others can extract the same information from a dataset. Energy, in contrast, can only be used once.

Just like the first industrial revolutions had a major impact on technology, economy and society, the same will likely apply for the current changes. It is arguably our information processing ability that is the basis of human dominance on this planet, and thus also of the major impact of humans on our planet. Since it is about information processing, the current revolution is thus potentially even more signficant than the first two industrial revolutions. We should strive to use these technologies well, to ensure they will contribute towards the solutions of humankind's and our planet's problems.
This extends from questions of ethical generation of energy (e.g., environmental concerns) to ethical generation of information (privacy, clickwork) all the way to how we are governed. In the beginnings of the information revolution, cybernetician Stafford Beer worked with Chile's Allende government to build cybernetic governance mechanisms \citep{Medina:2011:CRT:2086749}. In the current data-driven phase of this revolution, China is beginning to use machine learning to observe and incentivize citizens to behave in ways deemed beneficial \citep{chen2018transparent,daitoward}.
It is hard to predict where this development takes us --- this is science fiction at best, and the best science fiction may provide insightful thoughts on the topic.\footnote{Quoting from \citet{Asimov51}: {\em Hari Seldon [..] brought the science of psychohistory to its full development. [..] The individual human being is unpredictable, but the reactions of humans mobs, Seldon found, could be treated statistically.}}

\section{From Statistical to Causal Models}
\paragraph{Methods driven by independent and identically distributed (IID) data}
Our community has produced impressive successes with applications of machine learning to big data problems \citep{LeCBenHin15}. In these successes, there are multiple trends at work:
(1) we have massive amounts of data, often from simulations or large scale human labeling, (2) we use high capacity machine learning systems (i.e., complex function classes with many adjustable parameters), (3) we employ high performance computing systems, and finally (often ignored, but crucial when it comes to causality) (4) the problems are IID (independent and identically distributed).
The settings are typically either IID to begin with (e.g., image recognition using benchmark datasets), or they are artificially made IID, e.g.. by carefully collecting the right training set for a given application problem, or by methods such as DeepMind's ``experience replay'' \citep{mnih2015human} where a reinforcement learning agent stores observations in order to later permute them for the purpose of further training. 
For IID data, strong universal consistency results from statistical learning theory apply, guaranteeing convergence of a learning algorithm to the lowest achievable risk. Such algorithms do exist, for instance nearest neighbor classifiers and Support Vector Machines \citep{Vapnik98,SchSmo02,SteChr08}.
Seen in this light, it is not surprizing that we can indeed match or surpass human performance if given enough data. Machines often perform poorly, however, when faced with problems that violate the IID assumption yet seem trivial to humans. Vision systems can be grossly misled if an object that is normally recognized with high accuracy is placed in a context that {\em in the training set} may be negatively correlated with the presence of the object. For instance, such as system may fail to recognize a cow standing on the beach. Even more dramatically, the phenomenon of ``adversarial vulnerability'' highlights how even tiny but targeted violations of the IID assumption, generated by adding suitably chosen noise to images (imperceptible to humans), can lead to dangerous errors such as confusion of traffic signs. Recent years have seen a race between ``defense mechanisms'' and new attacks that appear shortly after and re-affirm the problem. Overall, it is fair to say that much of the current practice (of solving IID benchmark problems) as well as most theoretical results (about generalization in IID settings) fail to tackle the hard open problem of generalization across problems.

To further understand the way in which the IID assumption is problematic, let us consider a shopping example. Suppose Alice is looking for a laptop rucksack on the internet (i.e., a rucksack with a padded compartment for a laptop), and the web shop's recommendation system suggests that she should buy a laptop to go along with the rucksack. This seems odd because she probably already has a laptop, otherwise she would not be looking for the rucksack in the first place. In a way, the laptop is the cause and the rucksack is an effect. 
If I am told whether a customer has bought a laptop, it reduces my uncertainty about whether she also bought a laptop rucksack, and vice versa –-- and it does so by the same amount (the {\em mutual information}), so the directionality of cause and effect is lost.
It is present, however, in the physical mechanisms generating statistical dependence, for instance the mechanism that makes a customer want to buy a rucksack once she owns a laptop.
Recommending an item to buy constitutes an intervention in a system, taking us outside the IID setting. We no longer work with the observational distribution, but a distribution where certain variables or mechanisms have changed. This is the realm of causality.

\citet{Reichenbach1956}
clearly articulated the connection between causality and statistical dependence. He postulated the {\bf Common Cause Principle}: if two observables $X$ and $Y$ are statistically dependent, then there exists a variable $Z$ that causally influences both and explains all the dependence in the sense of making them independent when conditioned on $Z$. As a special case, this variable can coincide with $X$ or $Y$. 
Suppose that $X$ is the frequency of storks and $Y$ the human birth rate (in European countries, these have been reported to be correlated). If storks bring the babies, then the correct causal graph is $X\rightarrow Y$. If babies attract storks, it is $X \leftarrow Y$. If there is some other variable that causes both (such as economic development), we have $X \leftarrow Z \rightarrow Y$.

The crucial insight is that without additional assumptions, we cannot distinguish these three cases using observational data. The class of observational distributions over $X$ and $Y$ that can be realized by these models is the same in all three cases. A causal model thus contains genuinely more information than a statistical one.

Given that already the case where we have two observables is hard, one might wonder if the case of more observables is completely hopeless. Surprisingly, this is not the case: the problem in a certain sense becomes easier, and the reason for this is that in that case, there are nontrivial conditional independence properties \citep{Spohn78,Dawid79,GeiPea90} implied by causal structure. These can be described by using the language of causal graphs or structural causal models, merging probabilistic graphical models and the notion of interventions \citep{Pearl2009,Spirtes2000} best described using directed functional parent-child relationships rather than conditionals. While conceptually simple in hindsight, this constituted a major step in the understanding of causality, as later expressed by \citet[p. 104]{Pearl2009}:
\begin{quote}{\em We played around with the possibility of replacing the parents-child relationship $P(X_i|\PA_i)$ with its functional counterpart $X_i = f_i(\PA_i,U_i)$ and, suddenly, everything began to fall into place: We finally had a mathematical object to which we could attribute familiar properties of physical mechanisms instead of those slippery epistemic probabilities $P(X_i|\PA_i)$ with which we had been working so long in the study of Bayesian networks.}
\end{quote}

\paragraph{Structural causal models (SCMs)}
The SCM viewpoint is intuitive for those machine learning researchers who are more accustomed to thinking in terms of estimating functions rather than probability distributions. In it, we are given a set of {\bf observables} $X_1,\dots,X_n$ (modelled as random variables) associated with the vertices of a directed acyclic graph (DAG) $G$. We assume that each observable is the result of an assignment
\begin{equation}\label{eq:SA}
X_i := f_i (\PA_i, U_i),   ~~~~ (i=1,\dots,n),
\end{equation}
using a deterministic function $f_i$ depending on $X_i$'s parents in the graph (denoted by $\PA_i$) and on a stochastic {\em unexplained} variable $U_i$. Directed edges in the graph represent direct causation, since the parents are connected to $X_i$ by directed edges and through \eq{eq:SA} directly affect the assignment of $X_i$. The noise $U_i$ ensures that the overall object \eq{eq:SA} can represent a general conditional distribution $p(X_i|\PA_i)$, and the set of noises $U_1,\dots,U_n$ are assumed to be {\bf jointly independent}. If they were not, then by the Common Cause Principle there should be another variable that causes their dependence, and thus our model would not be {\em causally sufficient}.

If we specify the distributions of $U_1,\dots,U_n$, recursive application of \eq{eq:SA} allows us to compute the entailed observational joint distribution $p(X_1,\dots,X_n)$. This distribution has structural properties inherited from the graph \citep{Pearl2009,Lauritzen1996}: it satisfies the {\bf causal Markov condition} stating that conditioned on its parents, earch $X_j$ is independent of its non-descendants. Intuitively, we can think of the independent noises as ``information probes'' that spread through the graph (much like independent elements of gossip can spread through a social network). Their information gets entangled, manifesting itself in a footprint of conditional dependences rendering the possibility to infer aspects of the graph structure from observational data using independence testing. Like in the gossip analogy, the footprint may not be sufficiently characteristic to pin down a unique causal structure. In particular, it certainly is not if there are only two observables, since any nontrivial conditional independence statement requires at least three variables.

We have studied the two-variable problem over the last decade. We realized that it can be addressed by making additional assumptions, as not only the graph topology leaves a footprint in the observational distribution, but the functions $f_i$ do, too. This point is interesting for machine learning, where much attention is devoted to properties of function classes (e.g., priors or capacity measures), and we shall return to it below. Before doing so, we note two more aspects of \eq{eq:SA}. First, the SCM language makes it straightforward to formalize {\bf interventions} as operations that modify a subset of assignments \eq{eq:SA}, e.g., changing $U_i$, or setting $f_i$ (and thus $X_i$) to a constant \citep{Pearl2009,Spirtes2000}. Second, the graph structure along with the joint independence of the noises implies a canonical factorization of the joint distribution entailed by \eq{eq:SA} into causal conditionals that we will refer to as the {\bf causal (or disentangled) factorization},
\begin{equation}\label{eq:cf}
p(X_1,\dots,X_n) = \prod_{i=1}^n  p(X_i \mid \PA_i).
\end{equation}
While many other {\bf entangled factorizations} are possible, e.g., 
\begin{equation}\label{eq:non-cf}
p(X_1,\dots,X_n) = \prod_{i=1}^n p(X_i \mid X_{i+1},\dots,X_n),
\end{equation} 
Eq.~\eq{eq:cf} is the only one that decomposes the joint distribution into conditionals corresponding to the structural assignments \eq{eq:SA}. We think of these as the {\bf causal mechanisms} that are responsible for all statistical dependences among the observables. Accordingly, in contrast to \eq{eq:non-cf}, the disentangled factorization represents the joint distribution as a product of causal mechanisms.

The conceptual basis of statistical learning is a joint distribution $p(X_1,\dots,X_n)$ (where often one of the $X_i$ is a response variable denoted as $Y$), and we make assumptions about function classes used to approximate, say, a regression $\E (Y|X)$. {\bf Causal learning} considers a richer class of assumptions, and seeks to exploit the fact that the joint distribution possesses a causal factorization \eq{eq:cf}. It involves the causal conditionals $p(X_i \mid \PA_i)$ (i.e., the functions $f_i$ and the distribution of $U_i$ in \eq{eq:SA}), how these conditionals relate to each other, and interventions or changes that they admit. We shall return to this below.

\section{Levels of Causal Modelling}
Being trained in physics, I like to think of a set of coupled differential equations as the gold standard in modelling physical phenomena. It allows us to predict the future behavior of a system, to reason about the effect of interventions in the system, and --- by suitable averaging procedures --- to predict {\em statistical} dependences that are generated by coupled time evolutions.\footnote{Indeed, one could argue that indeed all statistical dependences in the world are due to such coupling.} It also allows us to gain insight in a system, explain its functioning, and in particular read off its causal structure:
consider the coupled set of differential equations
\begin{equation}\label{eq:ode}
\frac{d\x}{dt} = f(\x), \; \x \in \RR^d,
\end{equation}
with initial value $\x(t_0)=\x_0$. The Picard–Lindelöf theorem states that at least locally, if $f$ is Lipschitz, there exists a unique solution $\x(t)$. This implies in particular that the immediate future of $\x$ is implied by its past values.

If we formally write this in terms of infinitesimal differentials $dt$ and $d\x = \x(t+dt)-\x(t)$, we get:
\begin{equation}
\x(t+dt) = \x(t) + dt\cdot f(\x(t)).
\end{equation}
From this, we can ascertain which entries of the vector $\x(t)$ cause the future of others $\x(t+dt)$, i.e., the causal structure. 
This tells us that if we have a physical system that we can model using such an ordinary differential equation \eq{eq:ode}, solved for $\frac{d\x}{dt}$ (i.e., the derivative only appears on the left hand side), then its causal structure can be directly read off.

While a differential equation is a rather complete description of a system, a statistical model can be viewed as a much more superficial one. It usually does not talk about time; instead, it tells us how some of the variables allow prediction of others as long as experimental conditions do not change. E.g., if we drive a differential equation system with certain types of noise, or we average over time, then it may be the case that statistical dependences between components of $\x$ emerge, and those can then be exploited by machine learning. Such a model does not allows us to predict the effect of interventions; however, its strength is that it can often be learned from data, while a differential equation usually requires an intelligent human to come up with it.
Causal modelling lies in between these two extremes. It aims to provide understanding and predict the effect of interventions. Causal discovery and learning tries to arrive at such models in a data-driven way, using only weak assumptions.\footnote{It has been pointed out that this task is impossible without assumptions, but this is similar for the (easier) problems of machine learning from finite data. We \emph{always} need assumptions when we perform nontrivial inference from data.}
The overall situation is summarized in Table~\ref{t:taxonomy}, adapted from \citet{PetJanSch17}.

\begin{table}[tbh]
\caption{\label{t:taxonomy}A simple taxonomy of models. The most detailed model (top) is a mechanistic or physical one, usually in terms of differential equations. At the other end of the spectrum (bottom), we have a purely statistical model; this can be learned from data, but it often provides little insight beyond modeling associations between epiphenomena.
Causal models can be seen as descriptions that lie in between, abstracting away from physical realism while retaining the power to answer certain interventional or counterfactual questions. See also \citet{MooJanSch13} for a formal link between physical models and structural causal models.}
{\small
\begin{tabular}{ c || c | c | c | c | c }
Model& Predict in IID & Predict under distr.\ & Answer counter- & Obtain & Learn from  \\
 & setting & shift/intervention & factual questions & physical insight & data \\\hline\hline
Mechanistic/physical  & yes & yes & yes & yes & ? \\\hline
Structural causal  & yes & yes & yes & ? & ?\\\hline
Causal graphical & yes & yes & no & ? & ?\\\hline
Statistical & yes & no & no & no & yes
\end{tabular}}
\end{table}

\section{Independent Causal Mechanisms\label{sec:icm}}
\begin{figure}[tbh]
\begin{minipage}[c]{0.59\textwidth}
\begin{center}
\includegraphics[width=0.48\textwidth]{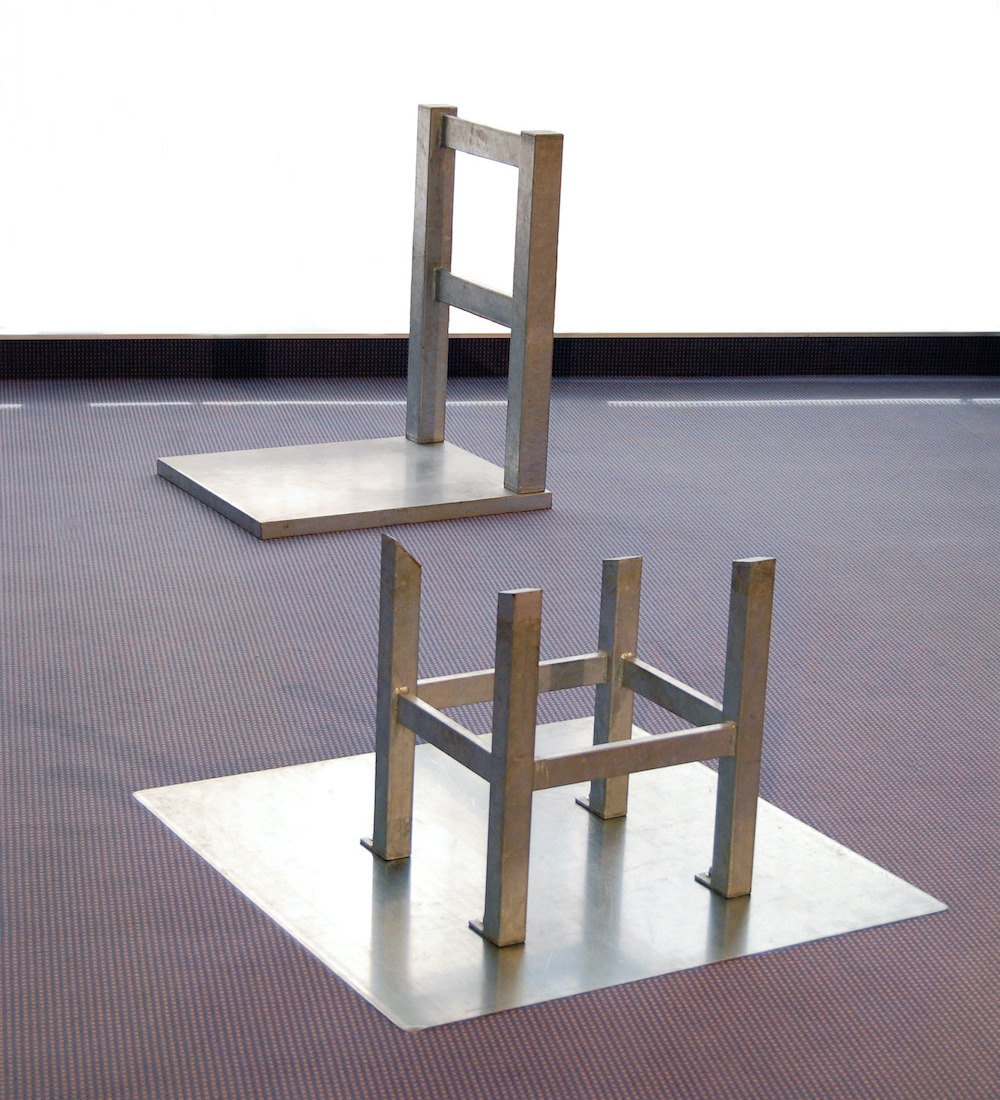}
\includegraphics[width=0.48\textwidth]{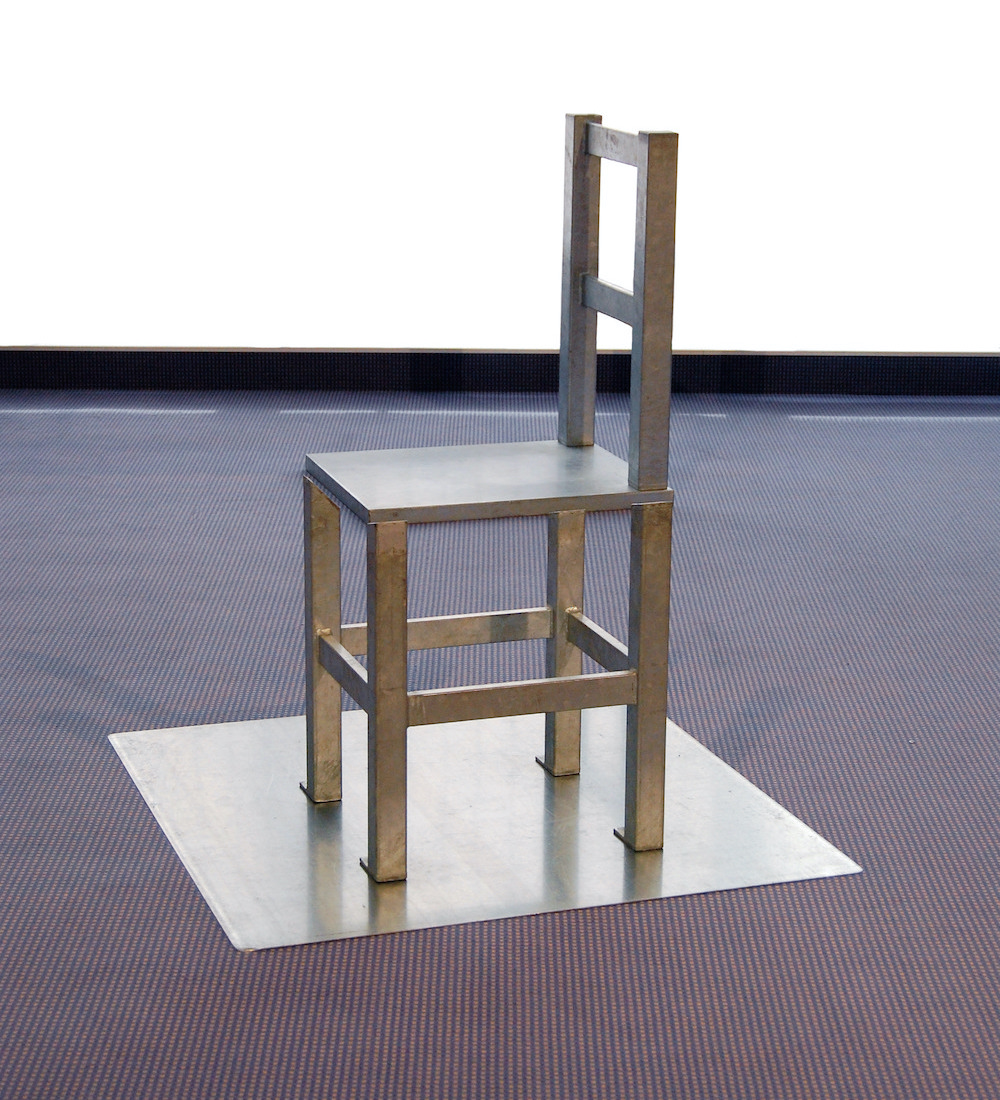}
\end{center}
\end{minipage}
\hfill
\begin{minipage}[c]{0.36\textwidth}
  \caption{{\em Beuchet chair}, made up of two separate objects that appear as a chair when viewed from a special vantage point violating the independence between object and perceptual process. (Image courtesy of Markus Elsholz, reprinted from \citet{PetJanSch17}.)\label{fig_BeuchetChair}}
  \end{minipage}
\end{figure}

We now return to the disentangled factorization \eq{eq:cf} of the joint distribution $p(X_1,\dots,X_n)$. This factorization according to the causal graph is always possible when the $U_i$ are independent, but we will now consider an additional notion of independence relating the factors in \eq{eq:cf} to one another. We can informally introduce it using an optical illusion known as the Beuchet Chair, shown in Figure~\ref{fig_BeuchetChair}. 

Whenever we perceive an object, our brain makes the assumption that the object and the mechanism by which the information contained in its light reaches our brain are {\bf independent}. We can violate this by looking at the object from an accidental viewpoint. If we do that, perception may go wrong: in the case of the Beuchet chair, we perceive the three-dimensional structure of a chair that in reality is not there.
The above independence assumption is useful because in practice, it holds most of the time, and our brain thus relies on objects being independent of our vantage point and the illumination. Likewise, there should not be accidental coincidences, 3D structures lining up in 2D, or shadow boundaries coinciding with texture boundaries. In vision research, this is called the generic viewpoint assumption. Likewise, if we move around the object, our vantage point changes, but we assume that the other variables of the overal generative process (e.g., lighting, object position and structure) are unaffected by that. This is an {\bf invariance} implied by the above independence, allowing us to infer 3D information even without stereo vision (``structure from motion''). An example of an extreme violation of this principle would be a head-mounted VR display tracking the position of a perceiver's head, and adjusting the display accordingly. Such a device can create the illusion of visual scenes that do not correspond to reality.

For another example, consider a dataset that consists of altitude $A$ and average annual temperature $T$ of weather stations \citep{PetJanSch17}. $A$ and $T$ are correlated, which we believe is due to the fact that the altitude has a causal effect on the temperature. Suppose we had two such datasets, one for Austria and one for Switzerland. The two joint distributions may be rather different, since the marginal distributions $p(A)$ over altitudes will differ. The conditionals $p(T|A)$, however, may be rather similar, since they characterize the physical mechanisms that generate temperature from altitude. However, this similarity is lost upon us if we only look at the overall joint distribution, without information about the causal structure $A\to T$. The causal factorization $p(A)p(T|A)$ will contain a component $p(T|A)$ that generalizes across countries, while the entangled factorization $p(T)p(A|T)$ will exhibit no such robustness.
Cum grano salis, the same applies when we consider interventions in a system. For a model to correctly predict the effect of interventions, it needs to be robust with respect to generalizing from an observational distribution to certain {\em interventional} distributions.

One can express the above insights as follows \citep{SchJanPetSgoetal12,PetJanSch17}:

\hypertarget{pri:im}{{\bf Independent Causal Mechanisms (ICM) Principle.}} \hspace{0.1mm}
{\em The causal generative process of a system's variables is composed of autonomous modules that do not inform or influence each other.

  In the probabilistic case, this means that the conditional distribution of each variable given its causes (i.e., its mechanism) does not inform or influence the other mechanisms. }

This principle subsumes several notions important to causality, including separate intervenability of causal variables, modularity and autonomy of subsystems, and invariance \citep{Pearl2009,PetJanSch17}.
If we have only two variables, it reduces to an independence between the cause distribution and the mechanism producing the effect distribution.

Applied to the causal factorization \eq{eq:cf}, the principle tells us that the factors should be independent in the sense that
\begin{itemize}
\item[(a)] changing (or intervening upon) one mechanism $p(X_i|\PA_i)$ does not change the other mechanisms $p(X_j|\PA_j)$ ($i\ne j$), and 
\item[(b)] knowing some other mechanisms $p(X_i|\PA_i)$ ($i\ne j$) does not give us information about a mechanism $p(X_j|\PA_j)$. 
\end{itemize}
Our notion of independence thus subsumes two aspects: the former pertaining to influence, and the latter to information.

We view any real-world distribution as a product of causal mechanisms. A change in such a distribution (e.g., when moving from one setting/domain to a related one) will always be due to changes in at least one of those mechanisms. Consistent with the independence principle, we hypothesize that smaller {\em changes tend to manifest themselves in a sparse or local way}, i.e., they should usually not affect all factors simultaneously. In contrast, if we consider a non-causal factorization, e.g., \eq{eq:non-cf}, then many terms will be affected simultaneously as we change one of the physical mechanisms responsible for a system's statistical dependences. Such a factorization may thus be called {\bf entangled}, a term that has gained popularity in machine learning \citep{1206.5538,1811.12359,Suter.1811.00007}.

The notion of invariant, autonomous, and independent mechanisms has appeared in various guises throughout the history of causality research.\footnote{Early work on this was done by \citet{Haavelmo1944}, stating the assumption that changing one of the structural assignments leaves the other ones invariant. 
\citet{Hoover06} attributes to Herb Simon the {\em invariance criterion}: the true causal order is the one that is invariant under the right sort of intervention.
\citet{Aldrich89} provides an overview of the historical development of these ideas in economics. He argues that the ``most basic question one can ask about a relation should be: How autonomous is it?'' \citep[][preface]{Frisch1948}. \citet{Pearl2009} discusses autonomy in detail, arguing that a causal mechanism remains invariant when other mechanisms are subjected to external influences. He points out that causal discovery methods may best work ``in longitudinal studies conducted under slightly varying conditions, where accidental independencies are destroyed and only structural independencies are preserved.'' 
Overviews are provided by \citet{Aldrich89,Hoover06,Pearl2009}, and \citet[Sec.~2.2]{PetJanSch17}.} Our contribution may be in unifying these notions with the idea of informational independence, and in showing that one can use rather general independence measures \citep{Steudel2010a}, a special case of which (algorithmic information) will be described below.

\paragraph{Measures of dependence of mechanisms}
Note that the dependence of two mechanisms $p(X_i|\PA_i)$ and $p(X_j|\PA_j)$ does not coincide with the statistical dependence of the random variables $X_i$ and $X_j$. Indeed, in a causal graph, many of the random variables will be dependent even if all the mechanisms are independent.

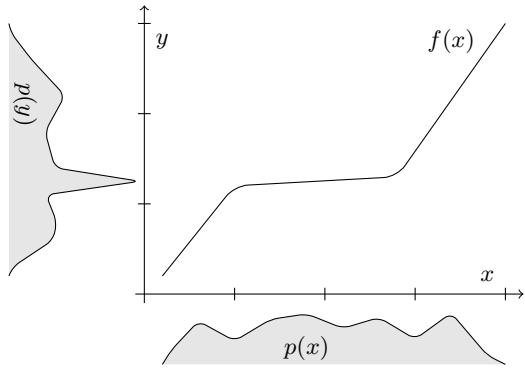
\begin{figure}
\begin{minipage}[c]{0.59\textwidth}
\centerline{
\begin{tikzpicture}[xscale=1.2, yscale=1.2]
\small
  \draw (0.2,2.8) node(y) {$y$};
  \draw (3.8,0.2) node(x) {$x$};
  \draw (3.4,2.8) node(x) {$f(x)$};
  \draw[->] (-0.1,0) -- (4.2,0); 
  \draw[->] (0,-0.1) -- (0,3.2); 
  \foreach \j in {1,2,3,4}{
  	\draw (\j,-0.07) -- (\j,0.07); 
  }
  \foreach \j in {1,2,3}{
  	\draw (-0.07,\j) -- (0.07,\j); 
  }  
\draw [rounded corners=5pt] (0.2,0.2) -- (1,1.2) -- (2.8,1.3) -- (4,3);
\filldraw [fill=gray!20, rounded corners=3pt] 
(0.2,-0.78) -- (0.3,-0.62) -- (0.6,-0.3) -- (1.0,-0.52) -- (1.4,-0.28) -- (1.8,-0.22) -- (2.2,-0.38) -- (2.6,-0.26) -- (3.0,-0.5) -- (3.4,-0.21) -- (3.8,-0.68) -- (4.0,-0.78);
  \draw (1.8,-0.6) node(px) {$p(x)$};
\filldraw [fill=gray!20, rounded corners=3pt] 
(-1.50,0.2) -- (-1.46,0.3) -- (-1.01,0.6) -- (-0.97,0.8) -- (-1.09,1.1) -- (-0.05,1.25) --  (-0.99,1.4) -- (-1.1,1.8) -- (-0.88,2.2) -- (-1.25,2.6) -- (-1.48,2.9) -- (-1.5,3.00);
  \draw (-1.32,2.1) node(py) [rotate=-90] {$p(y)$};
\end{tikzpicture}
$\qquad$
}
\end{minipage}
\begin{minipage}[c]{0.4\textwidth}
\caption{\label{fig:igci} (from \citet{PetJanSch17}) If $f$ and $p_x$ are chosen independently, then peaks of $p_Y$ tend to occur in regions where
$f$ has small slope and $f^{-1}$ has large slope. Thus $p_Y$ contains information about $f^{-1}$. 
}
\end{minipage}
\end{figure}

Intuitively speaking, the independent noise terms $U_i$ provide and parametrize the uncertainty contained in the fact that a mechanism $p(X_i|\PA_i)$ is non-deterministic, and thus ensure that each mechanism adds an independent element of uncertainty. I thus like to think of the \hyperlink{pri:im}{ICM Principle} as containing the independence of the unexplained noise terms in an SCM \eq{eq:SA} as a special case.\footnote{See also \citet{PetJanSch17}. Note that one can also implement the independence principle by assigning independent priors for the causal mechanisms. We can view \hyperlink{pri:im}{ICM} as a meta-level independence, akin to assumptions of time-invariance of the laws of physics \citep{Bohm}.}
However, it goes beyond this, as the following example illustrates. Consider two variables and structural assignments $X:= U$ and $Y:=f(X)$. I.e., the cause $X$ is a noise variable (with density $p_X$), and the effect $Y$ is a deterministic function of the cause. Let us moreover assume that the ranges of $X$ and $Y$ are both $[0,1]$, and $f$ is strictly monotonically increasing.
The principle of independent causal mechanisms then reduces to the independence of $p_X$ and $f$. Let us consider $p_x$ and the derivative $f'$ as random variables on the probability space $[0,1]$ with Lebesgue measure, and use their correlation as a measure of dependence of mechanisms.\footnote{Other dependence measures have been proposed for high-dimensional linear settings and time series by \citet{JanHoySch10,Shajarisales15,BesShaSchJan18,JanSch18b}, see also \citet{Janzing_NIPS2019}.} It can be shown that for $f\ne id$, independence of $p_X$ and $f'$ implies dependence between $p_Y$ and $(f^{-1})'$ (see Figure~\ref{fig:igci}). Other measures are possible and admit information-geometric interpretations. Intuitively, under the \hyperlink{pri:im}{ICM} assumption, the ``irregularity'' of the effect distribution becomes a sum of irregularity already present in the input distribution and irregularity introduced by the function, i.e., the irregularities of the two mechanisms add up rather than compensating each other, which would not be the case in the anti-causal direction (for details, see \citet{Janzingetal12}).

\paragraph{Algorithmic independence}
So far, I have discussed links between causal and statistical structures. The fundamental of the two is the causal structure, since it captures the physical mechanisms that generate statistical dependences in the first place. The statistical structure is an epiphenomenon that follows if we make the unexplained variables random.
It is awkward to talk about the (statistical) information contained in a mechanism, since deterministic functions in the generic case neither generate nor destroy information. This motivated us to devise an algorithmic model of causal structures in terms of Kolmogorov complexity \citep{JanSch10}. The Kolmogorov complexity (or algorithmic information) of a bit string is essentially the length of its shortest compression on a Turing machine, and thus a measure of its information content. Independence of mechanisms can be defined as vanishing mutual algorithmic information; i.e., two conditionals are considered independent if knowing (the shortest compression of) one does not help us achieve a shorter compression of the other one.

Algorithmic information theory provides a natural framework for non-statistical graphical models. Just like the latter are obtained from structural causal models by making the unexplained variables $U_i$ random, we obtain algorithmic graphical models by making the $U_i$ bit strings (jointly independent across nodes) and viewing the node $X_i$ as the output of a fixed Turing machine running the program $U_i$ on the input $\PA_i$. Similar to the statistical case, one can define a local causal Markov condition, a global one in terms of d-separation, and an additive decomposition of the joint Kolmogorov complexity in analogy to \eq{eq:cf}, and prove that they are implied by the structural causal model \citep{JanSch10}. What is elegant about this approach is that it shows that causality is not intrinsically bound to statistics, and that independence of noises and the independence of mechanisms now coincide since the independent programs play the role of the unexplained noise terms.

The assumption of algorithmically independent mechanisms has intriguing implications for physics, as it turns out to imply the second law of thermodynamics (i.e., the arrow of time) \citep{Janzing2016}. Consider a process where an incoming ordered beam of particles (the cause) is scattered by an object (the mechanism). Then the outgoing beam (the effect) contains information about the object. That is what makes vision and photography possible: photons contain information about the objects at which they have been scattered.
Now we know from physics that microscopically, the time evolution is reversible. Nevertheless, the photons contain information about the object only \emph{after} the scattering. Why is this the case, or in other words, why do photographs show the past rather than the future?

The reason is the independence principle, which we apply to initial state and system dynamics, postulating that the two are algorithmically independent, i.e., knowing one does not allow a shorter description of the other one. Then we can prove that the Kolmogorov complexity of the system's state is non-decreasing under the time evolution. If we view Kolmogorov complexity as a measure of entropy, this means that the entropy of the state can only stay constant or increase, amounting to the second law of thermodynamics and providing us with the thermodynamic arrow of time.

Note that this does not contradict microscopic irreversibility of the dynamics: the resulting state after time evolution is clearly \emph{not} independent of the system dynamic: it is precisely the state that when fed to the inverse dynamics would return us to the original state, i.e., the ordered particle beam. If we were able to freeze all particles and reverse their momenta, we could thus return to the original configuration without violating our version of the second law.

\section{Cause-Effect Discovery}
Let us return to the problem of causal discovery from observational data. Subject to suitable assumptions such as \emph{faithfulness} \citep{Spirtes2000}, one can sometimes recover aspects of the underlying graph from observations by performing conditional independence tests. However, there are several problems with this approach.
One is that in practice, our datasets are always finite, and conditional independence testing is a notoriously difficult problem, especially if conditioning sets are continuous and multi-dimensional. So while in principle, the conditional independences implied by the causal Markov condition hold true irrespective of the complexity of the functions appearing in an SCM, for finite datasets, conditional independence testing is hard without additional assumptions.\footnote{We had studied this for some time with Kacper Chwialkowski, Arthur Gretton, Dominik Janzing, Jonas Peters, and Ilya Tolstikhin; a formal result was obtained by \citet{1804.07203}.}
The other problem is that in the case of only two variables, the ternary concept of conditional independences collapses and the Markov condition thus has no nontrivial implications. 

It turns out that both problems can be addressed by making assumptions on function classes. This is typical for machine learning, where it is well-known that finite-sample generalization without assumptions on function classes is impossible. Specifically, although there are learning algorithms that are universally consistent, i.e., that approach minimal expected error in the infinite sample limit, for any functional dependence in the data, there are cases where this convergence is arbitrarily slow. So for a given sample size, it will depend on the problem being learned whether we achieve low expected error, and statistical learning theory provides probabilistic guarantees in terms of measures of complexity of function classes \citep{DevGyoLug96,Vapnik98}. 

Returning to causality, we provide an intuition why assumptions on the functions in an SCM should be necessary to learn about them from data. Consider a toy SCM with only two observables $X\to Y$. In this case, \eq{eq:SA} turns into
\begin{align}
X & = U \\
Y & = f(X, V) \label{eq:SA2}
\end{align}
with $U\independent V$.
Now think of $V$ acting as a random selector variable choosing from among a set of functions ${\cal F} = \{ f_v (x) \equiv f(x,v) \; | \; v \in \mbox{supp}(V)\}$. If $f(x,v)$ depends on $v$ in a non-smooth way, it should be hard to glean information about the SCM from a finite dataset, given that $V$ is not observed and it randomly switches between arbitrarily different $f_v$.\footnote{Suppose $X$ and $Y$ are binary, and $U,V$ are uniform Bernoulli variables, the latter selecting from ${\cal F} = \{ id, not\}$ (i.e., identity and negation). In this case, the entailed distribution for $Y$ is uniform, {\em independent} of $X$, even though we have $X\to Y$. We would be unable to discern $X\to Y$ from data.}
This motivates restricting the complexity with which $f$ depends on $V$. A natural restriction is to assume an {\bf additive noise model}
\begin{align}
X & = U \\
Y & = f(X) + V.
\end{align}
If $f$ in \eq{eq:SA2} depends smoothly on $V$, and if $V$ is relatively well concentrated, this can be motivated by a local Taylor expansion argument. It drastically reduces the effective size of the function class --- without such assumptions, the latter could depend exponentially on the cardinality of the support of $V$. 

Restrictions of function classes not only make it easier to learn functions from data, but it turns out that they can break the symmetry between cause and effect in the two-variable case: one can show that given a distribution over $X,Y$ generated by an additive noise model, one cannot fit an additive noise model in the opposite direction (i.e., with the roles of $X$ and $Y$ interchanged) \citep{Hoyer2008,Mooij2009,Peters2013anm,Kpotufe14,BauSchPet16}, cf.\ also the work of \citet{Sun2006}. This is subject to certain genericity assumptions, and notable exceptions include the case where $U,V$ are Gaussian and $f$ is linear. It generalizes results of \citet{Shimizu2006} for linear functions, and it can be generalized to include nonlinear rescalings \citep{Zhang2009}, loops \citep{Mooij11}, confounders \citep{Janzing2009uai}, and multi-variable settings \citep{Peters2011b}. We have collected a set of benchmark problems for cause-effect inference, and by now there is a number of methods that can detect causal direction better than chance \citep{Mooijetal16}, some of them building on the above Kolmogorov complexity model \citep{Vreeken}, and some directly learning to classify bivariate distributions into causal vs.\ anticausal \citep{LopMuaSchTol15}. This development has been championed by Isabelle Guyon whom (along with Andre Elisseeff) I had known from my previous work on kernel methods, and who had moved into causality through her interest in feature selection \citep{GuyonCausalFeatureSelection}.

Assumptions on function classes have thus helped address the cause-effect inference problem. They can also help address the other weakness of causal discovery methods based on conditional independence testing. Recent progress in (conditional) independence testing heavily relies on kernel function classes to represent probability distributions in reproducing kernel Hilbert spaces \citep{Gretton2005,Gretton2005JMLR,Fukumizu2008,Zhang2011uai,PfiBuhSchPet18,1804.02747}.

We have thus gathered some evidence that ideas from machine learning can help tackle causality problems that were previously considered hard. Equally intriguing, however, is the opposite direction: can causality help us improve machine learning?
Present-day machine learning (and thus also much of modern AI) is based on statistical modelling, but as these methods becomes pervasive, their limitations are becoming apparent. I will return to this after a short application interlude.


\section{Half-Sibling Regression and Exoplanet Detection}
The application described below builds on causal models inspired by additive noise models and the ICM assumption. By a stroke of luck, it enabled a recent breakthrough in astronomy, detailed at the end of the present section. 

Launched in 2009, NASA's Kepler space telescope initially observed 150000 stars over four years, in search of exoplanet transits. These are events where a planet partially occludes its host star, causing a slight decrease in brightness, often orders of magnitude smaller than the influence of instrument errors. When looking at stellar light curves with our collaborators at NYU, we noticed that not only were these light curves very noisy, but the noise structure was often shared across stars that were light years apart. Since that made direct interaction of the stars impossible, it was clear that the shared information was due to the instrument acting as a confounder. We thus devised a method that (a) predicts a given star of interest from a large set of other stars chosen such that their measurements contain no information about the star's astrophysical signal, and (b) removes that prediction in order to cancel the instrument's influence.\footnote{For events that are localized in time (such as exoplanet transits), we further argued that the same applies for suitably chosen past and future values of the star itself, which can thus also be used as predictors.} We referred to the method as ``half-sibling'' regression since target and predictors share a parent, namely the instrument.
The method recovers the random variable representing the desired signal almost surely (up to a constant offset), for an additive noise model, and subject to the assumption that the instrument's effect on the star is in principle predictable from the other stars \citep{Scholkopfetal16}.

Meanwhile, the Kepler spacecraft suffered a technical failure, which left it with only two functioning reaction wheels, insufficient for the precise spatial orientation required by the original Kepler mission. NASA decided to use the remaining fuel to make further observations, however the systematic error was significantly larger than before --- a godsend for our method designed to remove exactly these errors. 
We augmented it with models of exoplanet transits and an efficient way to search light curves, leading to the discovery of 36 planet candidates \citep{Foreman-Mackeyetal15}, of which 21 were
subsequently validated as bona fide exoplanets \citep{Montet_2015}.
Four years later, astronomers found traces of water in the atmosphere of the exoplanet K2-18b --- the first such discovery for an exoplanet in the habitable zone, i.e., allowing for liquid water \citep{1909.04642,Tsiaras}. The planet turned out to be one that had been first been detected in our work \citep[exoplanet candidate EPIC 201912552]{Foreman-Mackeyetal15}.

\section{Invariance, Robustness, and Semi-Supervised Learning}
Around 2009 or 2010, we started getting intrigued by how to use causality for machine learning. In particular, the ``neural net tank urban legend''\footnote{For a recent account, cf.\ \url{https://www.gwern.net/Tanks}} seemed to have something to say about the matter. In this story, a neural net is trained to classify tanks with high accuracy, but subsequently found to have succeeded by focusing on a feature (e.g., time of day or weather) that contained information about the type of tank only due to the data collection process. Such a system would exhibit no robustness when tested on new tanks whose images were taking under different circumstances. My hope was that a classifier incorporating causality could be made invariant with respect to this kind of changes, a topic that I had earlier worked on using non-causal methods \citep{ChaSch02}.
 We started to think about connections between causality and covariate shift, with the intuition that causal mechanisms should be invariant, and likewise any classifier building on learning these mechanisms. However, many machine learning classifiers were not using causal features as inputs, and indeed, we noticed that they more often seemed to solve anticausal problems, i.e., they used effect features to predict a cause. 

Our ideas relating to invariance matured during a number of discussions with Dominik, Jonas, Joris Mooij, Kun Zhang, Bob Williamson and others, from a departmental retreat in Ringberg in April 2010 to a Dagstuhl workshop in July 2011. 
The pressure to bring them to some conclusion was significantly stepped up when I received an invitation to deliver a Posner lecture at the Neural Information Processing Systems conference. At the time, I was involved in founding a new Max Planck Institute, and it was getting hard to carve out enough time to make progress.\footnote{Meanwhile, Google was stepping up their activities in AI, and I even forwent the chance to have a personal meeting with Larry Page to discuss this arranged by Sebastian Thrun.} Dominik and I thus decided to spend a week in a Black Forest holiday house to work on this full time, and during that week in November 2011, we completed a draft manuscript suitably named {\em invariant.tex}, submitted to the arXiv shortly after \citep{SchJanPetZha2011}. The paper argued that causal direction is crucial for certain machine learning problems, that robustness (invariance) to covariate shift is to be expected and transfer is easier for learning problems where we predict effect from cause, and it made a nontrivial prediction for semi-supervised learning. 

\paragraph{Semi-supervised learning (SSL)}
Suppose our underlying causal graph is $X\to Y$, and at the same time we are trying to learn a mapping $X\to Y$. The causal factorization \eq{eq:cf} for this case is 
\begin{equation}
p(X,Y) = p(X) p(Y|X).
\end{equation}
The \hyperlink{pri:im}{ICM Principle} posits that the modules in a joint distribution's causal decomposition do not inform or influence each other.
This means that in particular, $p(X)$ should contain no information about $p(Y|X)$, which implies that SSL should be futile, in as far as it is using additional information about $p(X)$ (from unlabelled data) to improve our estimate of $p(Y|X=x)$.
What about the opposite direction, is there hope that SSL should be possible in that case? It turns out the answer was yes, due to the work on cause-effect inference using independent causal mechanisms mentioned in Section~\ref{sec:icm}. This work was done with Povilas \citet{Daniusisetal10}.\footnote{Povilas was an original Erasmus intern visiting from Lithuania. If an experiment was successful, he would sometimes report this with a compact ``works.'' The project won him the best student paper prize at UAI.} It introduced a measure of dependence between the input and the conditional of output given input, and showed that if this dependence is zero in the causal direction, then it would be strictly positive in the opposite direction. Independence of cause and mechanism in the causal direction would thus imply that in the backward direction (i.e., for anticausal learning), the distribution of the input variable should contain information about the conditional of output given input, i.e., the quantity that machine learning is usually concerned with. I had previously worked on SSL \citep{ChaSchZie06}, and it was clear that this was exactly the kind of information that SSL required when trying to improve the estimate of output give input by using unlabelled inputs. We thus predicted that {\em SSL should be impossible for causal learning problems, but feasible otherwise}, in particular for anticausal ones.

I presented our analysis and the above prediction in the Posner lecture. Although a few activities relating to causality had been present at the conference during the years before, in particular a workshop in 2008 \citep{GuyonJS2010}, it is probably fair to say that the Posner lecture helped pave the way for causality to enter the machine learning mainstream. Judea, who must have been waiting for this development for some time, sent me a kind e-mail in March 2012, stating ``[...] I watched the video of your super-lecture at nips. A miracle.''

A subsequent meta-analysis of published SSL benchmark studies corroborated our prediction, was added to the arXiv report, and the paper was narrowly accepted for ICML \citep{SchJanPetSgoetal12}. We were intrigued with these results since we felt they provided some structural insight into {\em physical} properties of learning problems, thus going beyond the applications or methodological advances that machine learning studies usually provided. The line of work provided rather fruitful \citep{zhang_domain_2013,WeiSchBalGro14,zhang_multi-source_2015,GonZhaLiuTaoSch16,HuaZhaZhaSanGlySch17,1610.03263,1809.09337,1903.06256,1812.04597,1810.11953,1807.08479,1802.03916,Li_2018_ECCV,1707.06422,RojSchTurPet18}, and nicely complementary to studies of Elias Bareinboim and Judea \citep{Bareinboim2014,1503.01603}.
When Jonas moved to Z\"urich to complete and defend his Ph.D.\ in Statistics at ETH, he carried on with the invariance idea, leading to a thread of work in the statistics community exploiting invariance for causal discovery and other tasks \citep{Peters2015,1810.11776,1706.08576,1710.11469}.\footnote{Jonas also played a central role in spawning a thread of causality research in industry. In March 2011, Leon Bottou, working for Microsoft at the time, asked me if I could send him a strong causality student for an internship. Jonas was happy to take up the challenge, contributing to the work of \citet{Bottou2013}, an early use of causality to learn large scale interacting systems. Leon, one of the original leaders of the field of deep learning, has since taken a strong interest in causality \citep{LopNisChiSchBot17}.}

On the SSL side, subsequent developments include further theoretical analyses \citep[Section~5.1.2]{JanSch15,PetJanSch17} and a form of conditional SSL \citep{KugMeyLooSch19}. 
The view of SSL as exploiting dependences between a marginal $p(x)$ and a non-causal conditional $p(y|x)$ is consistent with the common assumptions employed to justify SSL \citep{ChaSchZie06}. The \emph{cluster assumption} asserts that the labelling function (which is a property of $p(y|x)$) should not change within clusters of $p(x)$. The \emph{low-density separation assumption} posits that the area where $p(y|x)$ takes the value of $0.5$ should have small $p(x)$; and the \emph{semi-supervised smoothness assumption}, applicable also to continuous outputs, states that if two points in a high-density region are close, then so should be the corresponding output values. Note, moreover, that some of the theoretical results in the field use assumptions well-known from causal graphs (even if they do not mention causality): the {\em co-training theorem} \citep{BluMit98} makes a statement about learnability from unlabelled data, and relies on an assumption of predictors being conditionally independent given the label, which we would normally expect if the predictors are (only) caused by the label, i.e., an anticausal setting. This is nicely consistent with the above findings.

\paragraph{Adversarial vulnerability}
One can hypothesize that causal direction should also have an influence on whether classifiers are vulnerable to \emph{adversarial attacks}. These attacks have recently become popular, and consist of minute changes to inputs, invisible to a human observer yet changing a classifier's output \citep{1312.6199}. 

This is related to causality in several ways. First, these attacks clearly constitute violations of the IID assumption that underlies predictive machine learning. If all we want to do is prediction in an IID setting, then statistical learning is fine.
In the adversarial setting, however, the modified test examples are not drawn from the same distribution as the training examples: they constitute interventions optimized to reveal the non-robustness of the (anticausal) $p(y|x)$.

The adversarial phenomen also shows that the kind of robustness current classifiers exhibit is rather different from the one a human exhibits. If we knew both robustness measures, we could try to maximize one while minimizing the other. Current methods can be viewed as crude approximations to this, effectively modeling the human's robustness as a mathematically simple set, say, an $l_p$ ball of radius $\epsilon>0$: they often try to find examples which lead to maximal changes in the classifier's output, subject to the constraint that they lie in an $l_p$ ball in the pixel metric. This also leads to procedures for adversarial training, which are similar in spirit to old methods for making classifiers invariant by training on ``virtual'' examples \citep[e.g.]{SchSmo02}.

Now consider a factorization of our model into components (cf.\ \eq{eq:non-cf}).
If the components correspond to causal mechanisms, then we expect a certain degreee of robustness, since these mechanisms are properties of nature. In particular, if we learn a classifier in the causal direction, this should be the case. One may thus hypothesize that for causal learning problems (predicing effect from cause), adversarial examples should be impossible, or at least harder to find \citep{Schoelkopf2017icml,KilParSch19arxiv}. Recent work supports this view: it was shown that a possible defense against adversarial attacks is to solve the anticausal classification problem by modeling the causal generative direction, a method which in vision is referred to as {\em analysis by synthesis} \citep{schott2018towards}.

More generally, also for graphs with more than two vertices, we can speculate that structures composed of autonomous modules, such as given by a causal factorization \eq{eq:cf}, should be relatively robust with respect to swapping out or modifying individual components. We shall return to this shortly.

Robustness should also play a role when studying {\em strategic behavior}, i.e., decisions or actions that take into account the actions of other agents (including AI agents). Consider a system that tries to predict the probability of successfully paying back a credit, based on a set of features. The set could include, for instance, the current debt of a person, as well as their address. To get a higher credit score, people could thus change their current debt (by paying it off), or they could change their address by moving to a more affluent neighborhood. The former probably has a positive causal impact on the probability of paying back; for the latter, this is less likely. We could thus build a scoring system that is more robust with respect to such strategic behavior by only using causal features as inputs \citep{1905.09239}.

\paragraph{Multi-task learning}
Suppose we want to build a system that can solve multiple tasks in multiple environments.
Such a model could employ the view of learning as compression. Learning a function $f$ mapping $x$ to $y$ based on a training set $(x_1,y_1),\dots,(x_n,y_n)$ can be viewed as conditional compression of $y$ given $x$. The idea is that we would like to find the most compact system that can recover $y_1,\dots,y_n$ given $x_1,\dots,x_n$.
Suppose Alice wants to communicate the labels to Bob, given that both know the inputs. First, they agree on a finite set of functions $\Fcal$ that they will use. Then Alice picks the best function from the set, and tells Bob which one it is (the number of bits required will depend on the size of the set, and possibly on prior probabilities agreed between Alice and Bob). In addition, she might have to tell him the indices $i$ of those inputs for which the function does not correctly classify $x_i$, i.e., for which $f(x_i)\ne y_i$. There is a trade-off between choosing a huge function class (in which case it will cost many bits to encode the index of the function) and allowing too many training errors (which need to be encoded separately). It turns out that this trade-off beautifully maps to standard VC bounds from statistical learning theory \citep{Vapnik95}.
One could imagine generalizing this to a multi-task setting: suppose we have multiple datasets, sampled from similar but not identical SCMs. If the SCMs share most of the components, then we could compress multiple datasets (sampled from multiple SCMs) by encoding the functions in the SCMs, and it is plausible that the correct structure (in the two-variables case, this would amount to the correct causal direction) should be the most compact one, since it would be one where many functions are shared across datasets, and thus need only be encoded once.


\paragraph{Reinforcement Learning}
The program to move statistical learning towards causal learning has links to reinforcement learning (RL), a sub-field of machine learning. RL used to be (and still often is) considered a field that has trouble with real-world high-dimensional data, one reason being that feedback in the form of a reinforcement signal is relatively sparse when compared to label information in supervised learning.
The DeepQ agent \citep{mnih2015human} yielded results that the community would not have considered possible at the time, yet it still has major weaknesses when compared to animate intelligence. Two major issues can be stated in terms of questions \citep{Schoelkopf15,Schoelkopf2017icml}:

{\em Question 1: why is RL on the original high-dimensional ATARI games harder than on downsampled versions?} For humans, reducing the resolution of a game screen would make the problem harder, yet this is exactly what was done to make the DeepQ system work. Animals likely have methods to identify objects (in computer game lingo, ``sprites'') by grouping pixels according to ``common fate'' (known from Gestalt psychology) or common response to intervention.
This question thus is related to the question of what constitutes an object, which concerns not only perception
but also concerns how we interact with the world. We can pick up one object, but not half an object.
Objects thus also correspond to modular structures that can be separately intervened upon or manipulated. The idea that objects are defined by their behavior under transformation is a profound one not only in psychology, but also in mathematics, cf.\ \citet{klein1872vergleichende,maclane:71}.

{\em Question 2: why is RL easier if we permute the replayed data?} As an agent moves about in the world, it influences the kind of data it gets to see, and thus the statistics change over time. This violates the IID assumption, and as mentioned earlier, the DeepQ agent stores and re-trains on past data (a process the authors liken to dreaming) in order to be able to employ standard IID function learning techniques. However, temporal order contains information that animate intelligence uses. Information is not only contained in temporal order, but also in the fact that slow changes of the statistics effectively create a multi-domain setting. Multi-domain data have been shown to help identify causal (and thus robust) features, and more generally in the search for causal structure, by looking for invariances \citep{PetJanSch17}.
This could enable RL agents to find robust components in their models that are likely to generalize to other parts of the state space. One way to do this is to employ model-based RL using SCMs, an approach which can help address a problem of confounding in RL where time-varying and time-invariant unobserved confounders influence both actions and rewards \citep{1812.10576}. In such an approach, nonstationarities would be a feature rather than a bug, and agents would actively seek out regions that are different from the known ones in order to challenge their existing model and understand which components are robust. This search can be viewed and potentially analyzed as a form of {\em intrinsic motivation,} a concept related to latent learning in Ethology that has been gaining traction in RL \citep{NIPS2004_2552}.

Finally, a large open area in causal learning is the connection to dynamics. While we may naively think that causality is always about time, most existing causal models do not (and need not) talk about time. For instance, returning to our example of altitude and temperature, there is an underlying temporal physical process that ensures that higher places tend to be colder. On the level of microscopic equations of motion for the involved particles, there is a clear causal structure (as described above, a differential equation specifies exactly which past values affect the current value of a variable). However, when we talk about the dependence or causality between altitude and temperature, we need not worry about the details of this temporal structure --- we are given a dataset where time does not appear, and we can reason about how that dataset would look if we were to intervene on temperature or altitude.
It is intriguing to think about how to build bridges between these different levels of description. Some progress has been made in deriving SCMs that describe the interventional behavior of a coupled system that is in an equilibrium state and perturbed in an ``adiabatic'' way \citep{MooJanSch13}, with generalizations to oscillatory systems \citep{RubBonMooSch18}. There is no fundamental reason why simple SCMs should be derivable in general. Rather, an SCM is a high-level abstraction of an underlying system of differential equations, and such an equation can only be derived if suitable high-level variables can be defined \citep{Rubensteinetal17}, which is probably the exception rather than the rule.


RL is closer to causality research than the machine learning mainstream in that it sometimes effectively directly estimates do-probabilities. E.g., on-policy learning estimates do-probabilities for the interventions specified by the policy (note that these may not be hard interventions if the policy depends on other variables). However, as soon as off-policy learning is considered, in particular in the batch (or observational) setting \citep{Lange2012}, issues of causality become subtle \citep{1812.10576,1805.12298}. Recent work devoted to the field between RL and causality includes \citep{Bareinboim2015,1703.07718,1812.10576,1811.06272,1901.08162,Bareinboim_NIPS2019}.

\section{Causal Representation Learning}
Traditional causal discovery and reasoning assumes that the units are random variables connected by a causal graph. Real-world observations, however, are usually not structured into those units to begin with, e.g., objects in images \citep{LopNisChiSchBot17}. The emerging field of causal representation learning hence strives to learn these variables from data, much like machine learning went beyond symbolic AI in not requiring that the symbols that algorithms manipulate be given a priori (cf.\ \citet{geffner}).
Defining objects or variables that are related by causal models can amount to coarse-graining of more detailed models of the world. Subject to appropriate conditions, structural models can arise from coarse-graining of microscopic models, including microscopic structural equation models \citep{Rubensteinetal17}, ordinary differential equations \citep{RubBonMooSch18}, and temporally aggregated time series \citep{Gongetal17}. Although every causal models in economics, medicine, or psychology uses variables that are abstractions of more elementary concepts, it is challenging to state general conditions under which coarse-grained variables admit causal models with well-defined interventions \citep{1512.07942,Rubensteinetal17}. 

 The task of identifying suitable units that admit causal models is challenging for both human and machine intelligence, but it aligns with the general goal of modern machine learning to learn meaningful representations for data, where meaningful can mean \emph{robust}, \emph{transferable}, \emph{interpretable}, \emph{explainable}, or \emph{fair} \citep{NIPS2017_6995,Kilbertusetal17,ZhaBar18}.
 To combine structural causal modeling \eq{eq:SA} and representation learning, we should strive to embed an SCM into larger machine learning models whose inputs and outputs may be high-dimensional and unstructured, but whose inner workings are at least partly governed by an SCM.
A way to do so is to realize the unexplained variables as (latent) noise variables in a generative model. Note, moreover, that there is a natural connection between SCMs and the modern generative models: they both use what has been called the {\em reparametrization trick} \citep{KinWel13}, consisting of making desired randomness an (exogenous) input to the model (in an SCM, these are the unexplained variables) rather than an intrinsic component.

\paragraph{Learning transferable  mechanisms}
An artificial or natural agent in a complex world is faced with limited resources. This concerns training data, i.e., we only have limited data for each individual task/domain, and thus need to find ways of pooling/re-using data, in stark contrast to the current industry practice of large-scale labelling work done by humans. It also concerns computational resources: animals have constraints on the size of their brains, and evolutionary neuroscience knows many examples where brain regions get re-purposed. Similar constraints on size and energy apply as ML methods get embedded in (small) physical devices that may be battery-powered.
Future AI models that robustly solve a range of problems in the real world will thus likely need to re-use components, which requires that the components are robust across tasks and environments \citep{SchJanLop16}. 
An elegant way to do this is to employ a modular structure that mirrors a corresponding modularity in the world. In other words, if the world is indeed modular, in the sense that different components of the world play roles across a range of environments, tasks, and settings, then it would be prudent for a model to employ corresponding modules \citep{RIMs}. For instance, if variations of natural lighting (the position of the sun, clouds, etc.) imply that the visual environment can appear in brightness conditions spanning several orders of magnitude, then visual processing algorithms in our nervous system should employ methods that can factor out these variations, rather than building separate sets of face recognizers, say, for every lighting condition. If our brain were to compensate for the lighting changes by a gain control mechanism, say, then this mechanism in itself need not have anything to do with the physical mechanisms bringing about brightness differences. It would, however, play a role in a modular structure that corresponds to the role the physical mechanisms play in the world's modular structure. This could produce a bias towards models that exhibit certain forms of structural isomorphism to a world that we cannot directly recognize, which would be rather intriguing, given that ultimately our brains do nothing but turn neuronal signals into other neuronal signals.

A sensible inductive bias to learn such models is to look for independent causal mechanisms \citep{Locatello_Mixture}, and competitive training can play a role in this: for a pattern recognition task, \citet{ParKilRojSch18} show that learning causal models that contain independent mechanisms helps in transferring modules across substantially different domains. In this work, handwritten characters are distorted by a set of unknown mechanisms including translations, noise, and contrast inversion. A neural network attempts to undo these transformations by means of a set of modules that over time specialize on one mechanism each. For any input, each module attempts to produce a corrected output, and a discriminator is used to tell which one performs best. The winning module gets trained by gradient descent to further improve its performance on that input. It is shown that the final system has learned mechanisms such as translation, inversion or denoising, and that these mechanisms transfer also to data from other distributions, such as Sanskrit characters.
This has recently been taken to the next step, embedding a set of dynamic modules into a recurrent neural network, coordinated by a so-called attention mechanism \citep{RIMs}. This allows learning modules whose dynamics operate independently much of the time but occasionally interact with each other.

\paragraph{Learning disentangled representations}
We have earlier discussed the \hyperlink{pri:im}{ICM Principle} implying both the independence of the SCM noise terms in \eq{eq:SA} and thus the feasibility of the disentangled representation
\begin{equation}\label{eq:cf2}
p(S_1,\dots,S_n) = \prod_{i=1}^n  p(S_i \mid \PA_i)
\end{equation}
as well as the property that the conditionals $ p(S_i \mid \PA_i)$ be independently manipulable and largely invariant across related problems. Suppose we seek to reconstruct such a {\bf disentangled representation using independent mechanisms} \eq{eq:cf2} from data, but the causal variables $S_i$ are not provided to us a priori. Rather, we are given (possibly high-dimensional) $X=(X_1,\dots,X_d)$ (below, we think of $X$ as an image with pixels $X_1,\dots,X_d$), from which we should construct causal variables $S_1,\dots,S_n$ ($n\ll d$) as well as mechanisms, cf.\ \eq{eq:SA},
\begin{equation}
S_i := f_i (\PA_i, U_i),   ~~~~ (i=1,\dots,n),
\end{equation}
modeling the causal relationships among the $S_i$.
To this end, as a first step, we can use an {\em encoder} $q:\R^d \to \R^n$ taking $X$ to a latent ``bottleneck'' representation comprising the unexplained noise variables  $U=(U_1,\dots,U_n)$.
The next step is the mapping $f(U)$ determined by the structural assignments $f_1,\dots,f_n$.\footnote{Note that for a DAG, recursive substitution of structural assignments reduces them to functions of the noise variables only. Using recurrent networks, cyclic systems may be dealt with.}
Finally, we apply a {\em decoder} $p:\R^n \to \R^d$. If $n$ is sufficiently large, the system can be trained using reconstruction error to satisfy $p \circ f \circ q\approx id$ on the observed images.\footnote{If the causal graph is known, the topology of a neural network implementing $f$ can be fixed accordingly; if not, the neural network decoder learns the composition $\tilde{p} = p\circ f$. In practice, one may not know $f$, and thus only learn an autoencoder $\tilde{p}\circ q$, where the causal graph effectively becomes an unspecified part of $\tilde{p}$. By choosing the network topology, one can ensure that each noise should only feed into {\em one} subsequent unit (using connections skipping layers), and that all DAGs can be learnt.}
To make it causal, we use the \hyperlink{pri:im}{ICM Principle}, i.e., we should make the $U_i$ statistically independent, and we should make the mechanisms independent.
This can be done by ensuring that they be invariant across problems, or that they can be independently intervened upon: if we manipulate some of them, they should thus still produce valid images, which could be trained using the discriminator of a generative adversarial network \citep{GAN}.

While we ideally manipulate causal variables or mechanisms, we discuss the special case of intervening upon the latent noise variables.\footnote{Interventions on the $S_i$ can be done accordingly, including the case of decoders without encoder (e.g., GANs).}
One way to intervene is to replace noise variables with the corresponding values computed from other input images, a procedure that has been referred to as hybridization by \citet{1812.03253}. In the extreme case, we can hybridize latent vectors where {\em each} component is computed from another training example. For an IID training set, these latent vectors have statistically independent components by construction.

In such an architecture, the encoder is an anticausal mapping that recognizes or reconstructs causal drivers in the world. These should be such that in terms of them, mechanisms can be formulated that are transferable (e.g., across tasks). 
The decoder establishes the connection between the low dimensional latent representation (of the noises driving the causal model) and the high dimensional world; this part constitutes a causal generative image model.
The \hyperlink{pri:im}{ICM} assumption implies that if the latent representation reconstructs the (noises driving the) true causal variables, then interventions on those noises (and the mechanisms driven by them) are permissible and lead to valid generation of image data.

\paragraph{Learning interventional world models and reasoning}
Modern representation learning excels at learning representations of data that preserve relevant statistical properties \citep{1206.5538,LeCBenHin15}. It does so, however, without taking into account causal properties of the variables, i.e., it does not care about the interventional properties of the variables it analyzes or reconstructs.
I expect that going forward, causality will play a major role in taking representation learning to the next level, moving beyond the representation of statistical dependence structures towards models that support intervention, planning, and reasoning, realizing Konrad Lorenz' notion of \emph{thinking} as \emph{acting in an imagined space}.
This ultimately requires the ability to reflect back on one's actions and envision alternative scenarios, possibly necessitating (the illusion of) free will \citep{Pearl2009forbes}. The biological function of self-consciousness may be related to the need for a variable representing oneself in one's Lorenzian {\em imagined space}, and free will may then be a means to communicate about actions taken by that variable, crucial for social and cultural learning, a topic which has not yet entered the stage of machine learning research although it is at the core of human intelligence \citep{Henrich}.

\section{Personal Notes and Conclusion}
My first conscious encounter with Judea Pearl was in 2001, at a symposium on the \emph{Interface of Computing Sciences and Statistics}.\footnote{\url{https://www.ics.uci.edu/~interfac/}} We both spoke at this symposium, and I recall his talk, formalizing an area of scientific inquiry that I had previously considered solidly part of the realm of philosophy. It stirred the same fascination that had attracted me to the research that I was doing at that time, in statistical learning theory and kernel methods. I had a background in mathematics and physics, had dabbled in neural networks, and was impressed when in 1994 I met Vladimir Vapnik who taught me a statistical theory underlying the philosophical problems of induction and generalization. Judea Pearl, another giant of our still young field of AI, seemed to be doing the same on a rather different but equally fascinating problem. Like Vladimir, Judea left me with a lasting impression as someone who has mastered not just technicalities, but has gained access to profound philosophical understanding.
%
With kernel methods and learning theory taking off, I did not manage to go into depth on causality at the time. I did follow some of the work in graphical models which became a staple in machine learning, and I knew that although most researchers shied away from presenting these models as causal, this interpretation existed and formed a conceptual motivation for that field. 

I was brought in touch with causality research the second time in 2004, by my study friend Dominik Janzing. He was at the time working on quantum information, and spoke about causality in a course he taught in Karlsruhe. The student Xiaohai Sun followed that lecture and convinced Dominik to start working with him on a project. Eventually, the question of a Ph.D.\ project came up, and Dominik (who felt his own field was too far from that) decided to ask me whether a joint supervision would make sense. At the time, Vladimir Vapnik was visiting my lab, and after a long conversation, he agreed this could be interesting (``you should decide if you want to play this game'' --- by his standards, a fairly enthusiastic endorsement). I decided to take the risk, Xiaohai became a student in my lab in Tübingen, and in 2007, Dominik joined us. We also recruited the student Jonas Peters, who had taken part in a summer course I had taught in 2006, as well as the postdocs Joris Mooij and Kun Zhang, both independently driven towards the problem of causality. With Andre Elisseeff and Steffen Lauritzen, Dominik and I wrote a proposal to organize a causality workshop in Dagstuhl. This workshop took place in 2009, and helped us become members of the causality community; it was where I first personally met Peter Spirtes. 

I feel fortunate to have had such a strong team of people to do this work (including many whom I did not mention by name above), and I believe we have made a contribution to modern causality research and especially its links to machine learning: both by using learning methods to develop data-driven causal methods, and by using inspiration from causality to better understand machine learning and develop new learning methods. In that respect, representation learning and disentanglement are intriguing fields. I recall a number of discussions with Yoshua Bengio when I was a member of the review panel and advisory board of the CIFAR program. He was driven by the goal to disentangle the underlying factors of variation in data using deep learning, and I was arguing that this is a causal question. Our opinions have since then converged, and research has started to appear that combines both fields \citep{1901.10912,RIMs,1811.12359,Suter.1811.00007,1711.08936}.

All this is still in its infancy, and the above account is personal and thus biased --- I apologize for any omissions. With the current hype around machine learning, there is much to say in favor of some humility towards what machine learning can do, and thus towards the current state of AI --- the hard problems have not been solved yet, making basic research in this field all the more exciting.

\paragraph{Acknowledgments}
Many thanks to all past and present members of the T\"ubingen causality team, without whose work and insights this article would not exist, in particular to Dominik Janzing and Chaochao Lu who read a version of the manuscript. The text has also benefitted from discussions with Elias Bareinboim, Yoshua Bengio, Christoph Bohle, Leon Bottou, Anirudh Goyal, Isabelle Guyon, Judea Pearl, and Vladimir Vapnik.  Wouter van Amsterdam and Julius von K\"ugelgen have pointed out typos which have been corrected in this second version.

\bibliographystyle{mcp-acm}  
{\small
\bibliography{references}
}

\end{document}